\documentclass[pdflatex,sn-mathphys]{sn-jnl}

\jyear{2021}%

\usepackage{algorithm}
\usepackage[]{algpseudocode}
\usepackage{amsthm,amssymb,amsmath}

\theoremstyle{thmstyleone}%
\newtheorem{theorem}{Theorem}
\theoremstyle{thmstylethree}%

\newtheorem{definition}{Definition}%
\newtheorem{notation}{Notation}

\usepackage{listings}
\usepackage{xcolor}
\usepackage{soul}

\colorlet{punct}{red!60!black}
\definecolor{background}{HTML}{EEEEEE}
\definecolor{delim}{RGB}{20,105,176}
\colorlet{numb}{magenta!60!black}

\lstdefinelanguage{json}{
    basicstyle=\normalfont\ttfamily,
    numbers=left,
    numberstyle=\scriptsize,
    stepnumber=1,
    numbersep=8pt,
    showstringspaces=false,
    breaklines=true,
    frame=lines,
    backgroundcolor=\color{background},
    literate=
     *{0}{{{\color{numb}0}}}{1}
      {1}{{{\color{numb}1}}}{1}
      {2}{{{\color{numb}2}}}{1}
      {3}{{{\color{numb}3}}}{1}
      {4}{{{\color{numb}4}}}{1}
      {5}{{{\color{numb}5}}}{1}
      {6}{{{\color{numb}6}}}{1}
      {7}{{{\color{numb}7}}}{1}
      {8}{{{\color{numb}8}}}{1}
      {9}{{{\color{numb}9}}}{1}
      {:}{{{\color{punct}{:}}}}{1}
      {,}{{{\color{punct}{,}}}}{1}
      {\{}{{{\color{delim}{\{}}}}{1}
      {\}}{{{\color{delim}{\}}}}}{1}
      {[}{{{\color{delim}{[}}}}{1}
      {]}{{{\color{delim}{]}}}}{1},
}

\makeatletter
\newsavebox\myboxA
\newsavebox\myboxB
\newlength\mylenA

\newcommand*\xoverline[2][0.75]{%
    \sbox{\myboxA}{$\m@th#2$}%
    \setbox\myboxB\null
    \ht\myboxB=\ht\myboxA%
    \dp\myboxB=\dp\myboxA%
    \wd\myboxB=#1\wd\myboxA
    \sbox\myboxB{$\m@th\overline{\copy\myboxB}$}
    \setlength\mylenA{\the\wd\myboxA}
    \addtolength\mylenA{-\the\wd\myboxB}%
    \ifdim\wd\myboxB<\wd\myboxA%
       \rlap{\hskip 0.5\mylenA\usebox\myboxB}{\usebox\myboxA}%
    \else
        \hskip -0.5\mylenA\rlap{\usebox\myboxA}{\hskip 0.5\mylenA\usebox\myboxB}%
    \fi}
\makeatother

\begin{document}

\title[Linear Temporal Public Announcement Logic]{Linear Temporal Public Announcement Logic: a new perspective for reasoning about the knowledge of multi-classifiers}


\author[1]{\fnm{Amirhoshang} \sur{Hoseinpour Dehkordi}}\email{amir.hoseinpour@ipm.ir}

\author*[2]{\fnm{Majid} \sur{Alizadeh}}\email{majidalizadeh@ut.ac.ir}

\author[3]{\fnm{Ali} \sur{Movaghar}}\email{movaghar@sharif.edu}

\affil*[1]{\orgdiv{School of Computer Science}, \orgname{Institute for Research in Fundamental Sciences}, \orgaddress{
\postcode{19538-33511}, \state{Tehran}, \country{Iran}}}

\affil[2]{\orgdiv{School of Mathematics,
Statistics and Computer Science, College of Science}, \orgname{University of Tehran}, \orgaddress{
\postcode{14155-6455}, \state{Tehran}, \country{Iran}}}

\affil[3]{\orgdiv{Department of Computer Engineering}, \orgname{Sharif University of Technology}, \orgaddress{
\postcode{11155-9517}, \state{Tehran}, \country{Iran}}}


\abstract In this note, a formal transition system model called LTPAL to extract knowledge in a classification process is suggested. The model combines the Public Announcement Logic (PAL) and the Linear Temporal Logic (LTL). In the model, first, we consider classifiers, which capture single-framed data. Next, we took classifiers for data-stream data input into consideration. Finally, we formalize natural language properties in LTPAL with a video-stream object detection sample. 


\maketitle

\section{Introduction}
Nowadays, classification is inseparable from real-world applications. The presence of errors increase the necessity for better interpreting and understanding such algorithms.  
Due to the probabilistic nature of most of these approaches, the reasoning about results remains complicated. 
Previously, a model was developed by Dehkordi \textit{et al.} to verify properties for classifiers. In which, the analysis of the input data and external information in a multi-agent scenario was considered \cite{HoseinpourDehkordi2020}. The proposed method works fine for single-framed input data; however, it suffers from not being able to be used appropriately for semi-continuous data streams (i.e., video streams). More progress is also possible by finding the similarity of vicinity frames, as the information changes in such data streams are considered minor \cite{irani2001semi}. The process of data understanding can be divided into several steps. The first and foremost one is the object recognition from the input data, for which there are many solutions, often with great performances, \cite{redmon2016you}. The next step is knowledge extraction to understand the knowledge acquired in the initial step (see \cite{plaza2007logics}), where objects are transformed into symbols. In this step, knowledge can be shared and aggregated via a modal logic approach \cite{HoseinpourDehkordi2020}. Subsequently, the given rules, which are fed into the system, should be interpreted. 
This can be achieved by employing predefined protocols (i.e., ``cat is an animal'' or ``bat and pigeon could not have appeared in a single data frame'' as predefined protocols). The final step is to satisfy the specified formula, employing a collected set of knowledge. The last three steps will be highlighted further in this study.

\subsection{Related Works}
The reasoning is one of the most studied and accepted dimensions of AI (see \cite{winston1992artificial}, \cite{russell2002artificial}, \cite{nilsson2009quest}, \cite{goldman1992probabilistic}). Furthermore, it inspired researchers to target Visual data understanding, question answering, and natural language queries \cite{gao2017tall}, \cite{anne2017localizing}.
For instance, a question-answering (QA) reasoning system is developed by Bauer \textit{et al.}, where an algorithm was introduced to select the best paths of commonsense knowledge to get the whole inference required for QA \cite{bauer-etal-2018-commonsense}. The method works fine for specific applications, but it did not provide a general solution for QA problems. 
Based on the systematic analysis of popular knowledge resources and the knowledge-integration methods, Ma \textit{et al.}  developed a modelling solution \cite{ma2019towards}. The method, called ``non-extractive commonsense QA'', employs  ConceptNet (Speer \textit{et al.} \cite{speer2017conceptnet}) and the most recently introduced ATOMIC (Sap \textit{et al.} \cite{sap2019atomic}). More recently, based on human intelligence, the ``CoLlision Events for Video REpresentation and Reasoning'' method (CLEVRER) was developed by Yi and coworkers \cite{yi2019clevrer}, which is a reasoning system for video streams. 
In that method, unlike former studies, ``causal structures'' are also taken into account. Moreover, this model was able to answer four main varieties of questions: \textit{descriptive} (e.g., ``what color''), \textit{explanatory} (e.g., ``what’s responsible for''), \textit{predictive} (e.g., ``what will happen next''), and \textit{counterfactual} (e.g., ``what if''). 
In that model, the solution for extracting features from the video frames is ResNet-50, a well-known classifier \cite{he2016deep}. Thanks to the characteristics of the reasoning systems and the power of temporal logic in formalizing natural language, the model developed by Fong \textit{et al.} gets more interested in reasoning systems\cite{fong2019temporal}.
From another perspective, the Metric Temporal Logic (MTL), an extension of the Linear Temporal Logic (LTL), was examined to handle the stochastic state information \cite{de2019approximate}. Concerning typical AI-based problems, the employment of modal logic would be attractive. Recently, Dehkordi \textit{et al.} developed an epistemic-based model for the extraction and aggregation of knowledge from classifiers \cite{HoseinpourDehkordi2020}. The model works perfectly for single framed data; however, this model did not address data stream inputs.
\subsection{Motivation}
In the model introduced by Dehkordi \textit{et al.} \cite{HoseinpourDehkordi2020}, algorithms \ref{alg:CKC}, \ref{alg:MASKA}, \ref{alg:MASKS} were developed for knowledge sharing modelling in a multi-classifier scenario. In a network of trusted classifiers (which share their knowledge correctly), the process of knowledge extraction from one input for an agent is developed in algorithm \ref{alg:CKC}. In other words, all output classes of the input, presented in the neighbourhood set of $\eta_\epsilon(x)$ (here referred to as $\eta(x)$),  were collected. $\eta(x)$ is a set of inputs in $\epsilon$ vicinity (based on distance function $d(x_0,x)$). Here, the classifier function $\mathcal{N}(x)$ yields the respecting output classes by $\eta(x)$ input. Consequently, the knowledge of $\mathcal{N}$ -which contains the set of output classes of $\mathcal{N}(x)$- is represented by $\mathcal{K}$ (as the knowledge set of the classifier). The output is robust if the size of the set of output classes of $\mathcal{N}(x)$ is one. Therefore, from the agent's perspective, the output concludes the robustness of $x$. An agent's perspective is knowledge, which is accessible by that agent. 
\begin{algorithm}
Let $\mathcal{N}(x)=c$ be the function of the considered a classifier and $c$ be the result class\;
\caption{The Classifier Knowledge Calculator (CKC) function shall calculate the knowledge produced by a classifier for an input point, using neighbourhood function and manipulation function }
\label{alg:CKC}
\begin{algorithmic}[1]
\Function{CKC}{$\mathcal{N},x_0, \eta, \mu$}\\
	\Comment{$\mathcal{N},x_0, \eta, \mu$ are a classifier, an input point, neighbourhood function, and manipulation function respectively}
	\State $\mathcal{K} \gets  \emptyset$ 
	\ForAll{$x \in \eta(x_0) \cup \mu(x_0)$}
		\State $c \gets  \mathcal{N}(x)$ \Comment{$c$ represents the respective possible world}
		\If{$c \notin \mathcal{K}$}
			\State Add $c$ to $\mathcal{K}$ set
		\EndIf
	\EndFor
	\If{$\mid \mathcal{K}\mid  = 1$}
		\State \textbf{return} 1, $\mathcal{K}$
	\EndIf
	\State \textbf{return} 0, $\mathcal{K}$
\EndFunction
\end{algorithmic}
\end{algorithm}

Algorithm \ref{alg:MASKA} aggregated all knowledge of agents (i.e., the possible worlds) from the perspective of each agent. After finding the intersection, if it contains just one class, the input is verified. Here, if the output is an empty set, the MAS's knowledge is inconsistent. In the case of more than one output class, agents are not capable of concluding the answer. However, they know the answer is among these classes. 
\begin{algorithm}
\caption{The MAS Knowledge Aggregator (MASKA) function is going to aggregate the knowledge that is produced by a group of classifiers, for each input point using a neighbourhood function and manipulation function. }\label{alg:MASKA}
\begin{algorithmic}[1]
\Function{MASKA}{$\mathcal{N_S},x_0, \eta, \mu$}\\
    \Comment{$\mathcal{N_S},x_0, \eta, \mu$ are a group of classifiers, an input point, neighbourhood function, and manipulation function respectively}
	\State $\mathcal{K_S} \gets  set~of~all~output~classes$ 
	\ForAll{$\mathcal{N} \in \mathcal{N_S}$}
		\State $is\_Robust$, $\mathcal{K} \gets  $ CKC $(\mathcal{N},x_0, \eta, \mu)$ 
		\State $\mathcal{K_S} \gets \mathcal{K_S} \cap \mathcal{K} $ 
		\If{$\mathcal{K_S} = \emptyset$}
			\State \textbf{return} 0, $\emptyset$
		\EndIf
	\EndFor
	\If{$\mid \mathcal{K_S}\mid  = 1$}
		\State \textbf{return} 1, $\mathcal{K_S}$
	\EndIf
	\State \textbf{return} 0, $\mathcal{K_S}$
\EndFunction
\end{algorithmic}
\end{algorithm}

In algorithm \ref{alg:MASKS}, all captured knowledge is aggregated, and the reasoning is satisfied. 

\begin{algorithm}
\caption{The MAS Knowledge Sharing (MASKS) function shall aggregate the knowledge that is produced by an external system }\label{alg:MASKS}
\begin{algorithmic}[1]
\Function{MASKS}{$\mathcal{N_S},x_0, \eta, \mu$}\\
    \Comment{$\mathcal{N_S},x_0, \eta, \mu$ are a group of classifiers, an input point, neighbourhood function, and manipulation function respectively}
	\State $\mathcal{K_S} \gets  \emptyset$ 
	\ForAll{$\mathcal{N} \in \mathcal{N_S}$}
		\State $is\_Robust$, $\mathcal{K} \gets  $CKC$(\mathcal{N},x_0, \eta, \mu)$ 
		\State $\mathcal{K_S} \gets \mathcal{K_S} \cap \mathcal{K} $ 
		\If{$\mathcal{K_S} = \emptyset$}
			\State \textbf{return} 0, $\emptyset$
		\EndIf 
	\EndFor
	\If{$\mid \mathcal{K_S}\mid  = 1$} \\ \Comment{Check whether the knowledge can verify the input, or if more knowledge is needed}
		\State \textbf{return} 1, $\mathcal{K_S}$
	\EndIf
	\ForAll{$\mathcal{M} \in $ All knowledge sources}
		\State $is\_Robust$, $\mathcal{K} \gets  $ Announced knowledge\\
		\Comment{The external knowledge must be written in the DEL formula, where possible worlds are ones in which the formula is satisfied.}
		\State $\mathcal{K_S} \gets \mathcal{K_S} \cap \mathcal{K} $ 
		\If{$\mathcal{K_S} = \emptyset$}
			\State \textbf{return} 0, $\emptyset$
		\EndIf
	\EndFor
	\If{$\mid \mathcal{K_S}\mid  = 1$}
		\State \textbf{return} 1, $\mathcal{K_S}$
	\EndIf
	\State \textbf{return} 0, $\mathcal{K_S}$
\EndFunction
\end{algorithmic}
\end{algorithm}

Afterwards, this knowledge of single-framed data is modelled in a multi-agent scenario (formal definitions will be presented in section \ref{sec:PublicAnnouncementLogic}). 

This study aims to extend the expressiveness of the model developed by Dehkordi \textit{et al.} \cite{HoseinpourDehkordi2020} to present semi-continuous and multi-framed data. Besides, a flexible reasoning system is suggested to render the knowledge achieved by classifiers. Here, the model is applied to the existed classifiers to translate the information attained from the results. Furthermore, the ability of the model to handle multi-knowledge flow in a multi-classifier scenario will be assessed. 
For this purpose, various collaboration scenarios of classifiers (i.e., the investigation of knowledge of each agent, a group of agents, the knowledge distributed between agents, etc.) would be studied. The focus is on verifying the investigated formulas proposed for such a statement using the definition of ``verified formula'' mentioned in Dehkordi \textit{et al.} \cite{HoseinpourDehkordi2020}.
Certain questions (in human language) are translated to the developed temporal formula (as properties) to check the adaption and applicability of the proposed model. By checking the satisfiability of the temporal formula, the answers will be presented. 


 This note  has been structured in the following manner:

In section \ref{sec:ProblemDefinition}, the outline of the problem and the structure of the model is proposed. Then, formal verification of classifiers are defined in section \ref{sec:FormalverificationofClassifiers}. A PAL model is defined in section \ref{sec:PublicAnnouncementLogic} to capture the information from single-framed data. Finally, in part \ref{sec:LTPAL}, an extension of the model (called "Linear Temporal Public Announcement Logic", LTPAL) is defined by combining PAL with LTL. LTPAL is applied to capture semi-continuous data of multi-agent systems. In order to create a more convenient approach for defining verification's property, a procedure will be suggested to formalise natural language in LTPAL -which can be served as an application.

\section{Problem Definition}
\label{sec:ProblemDefinition}
Classification is a process of assigning a label (from a label set) to an input (vector). The solution comes with uncertainties in most classification approaches (i.e., neural networks). There are approaches in which multiple answers would be provided if the answer does not meet the minimum requirements (i.e., low probability or near boundary answers). Using the collection of answers, we could create a knowledge set. The knowledge set shows what are the possible answers using applied classifiers.
In this study, we will develop a model to extract knowledge in a classification process. Then, we consider classifiers, which capture single-framed data. Next, we took classifiers for data-stream data input into consideration. Finally, the developed model will be applied to two scenarios: question-answering systems and video-stream object detection.  

Note that, a set of agents (classifiers) also could refer to a group of agents (classifiers), so the phrase set and group could be used interchangeably.
Herein, we will develop a reasoning model on the top of a preexisted intelligent system, to reason about the knowledge, in the following steps:
\begin{enumerate}
    \item Classification is a process of getting an input vector of size x and assigning an element c from the set C to the vector x (c is the answer of x), in which c is the output class, and C is the set of all possible classes. First of all, we are going to collect all the intelligent system's possible answers. 
    Therefore, because of the statistical nature of most classifiers and the fact that the accuracy is based on the architecture and training set, this kind of classifier could not guarantee complete accuracy. To overcome this problem, Huang \textit{et al.} \cite{Huang0} presented an approach to defining robustness for each input (point-wise verification) in such models. In this approach, for each input $x$, a set of input points created by an arbitrary function $\eta(x)$ would be fed into a classifier; if all answers converged into one class, the input point is named robust for the classifier. More precisely, Dehkordi \textit{et al.} \cite{HoseinpourDehkordi2020} developed a multi-agent epistemic logic model to find all possible outcomes of multi-classifiers, together with defining verification of properties using the Kripke model. By applying this step, the outcome of such classifiers would be a Kripke model that carries a set of all possible outputs (i.e., possible worlds).
    In this study, if the size of the set of all possible worlds equals one, the model would be verified the input concerning the defined properties.
    \item Generally, the set of knowledge is directly driven by classifiers from the possible worlds. This set of knowledge for input and a system of intelligent agents contains all possible output classes, yet from a real-world perspective, each class of objects would contain more information (i.e., ontology rules).  To reflect this, an intelligent agent should understand the object's category with sub-categories fed in as input rules. In this model step, predefined inferences would be extracted in a unified and formal manner for each possible object class.
    \item The time-series data could not be considered until this step. To consider them, we developed a combination of temporal logic transition systems Gerth \textit{et al.} \cite{gerth1995simple} with the developed epistemic logic model. Let each frame of time-series data as an input. After executing the above steps on all frames, we have a set of extracted knowledge (in the Kripke model) for each single-framed data of the input time-series data. By placing the Kripke models in hierarchical order, a transition system could be created. This leads us to extract all possible sequences (named execution path) of knowledge represented in time-series data. 
    \item Then, we will develop an approach to use our model for a question-answering application. Accordingly, to get an outcome from the developed model, it should provide answers to questions. First, using preexisted approaches, natural language statements will be translated into temporal formulas. This formula could be inferred as a property to be verified in a verification problem. Then, each formula will be investigated in corresponded paths (respecting the type of answer, which we will discuss in future sections, it could be all or some paths). Finally, the questions related to a satisfying formula will be answered ``true'' and others ``false''. 
    \item The verification of property in this transition system is also defined by modifying the verification definition in Huang \textit{et al.} \cite{Huang0} and Dehkordi \textit{et al.} \cite{HoseinpourDehkordi2020}. Accordingly, the situation of a formula (equivalently a property) would be determined. In other words, it is checked whether it is a verified formula (verified in all paths), possible formula (verified in at least one path), or if adding some missing information (by adding information to the system it could be verified), it would be a possible/verified formula.
\end{enumerate}

Here, we will do the mentioned steps as an example of a video stream and classifiers. In this case, classifiers could determine objects appearing during the video. First, all possible outcomes for each image known as a data-frame will be extracted by the model developed in section \ref{sec:PublicAnnouncementLogic}. Then, by collecting all input rules of objects (i.e., an input rule could be a Sub-feature such as ``elephants'' are ``animals'' so, in each possible situation an ``elephant'' detected, we infer that an ``animal'' is also detected), each possible outcome will be enhanced for each image in the video. Next, the transition system would be made by placing the model of all possible images in a hierarchical sequencing (see section \ref{sec:LTPAL}). The Kripke model was created before this step. Consequently, and as an application, if any question was asked about the video, it would be investigated by translating questions into Linear Temporal Public Announcement Logic (LTPAL). The detail of the question-answering system will be shown in section \ref{sec:QALTPAL}. After that, for the sake of the performance, we will calculate the probability of each transition relation, following that, finding the most probable execution path. 

\begin{figure*}
  \centering
  \includegraphics[width=1.0\textwidth]{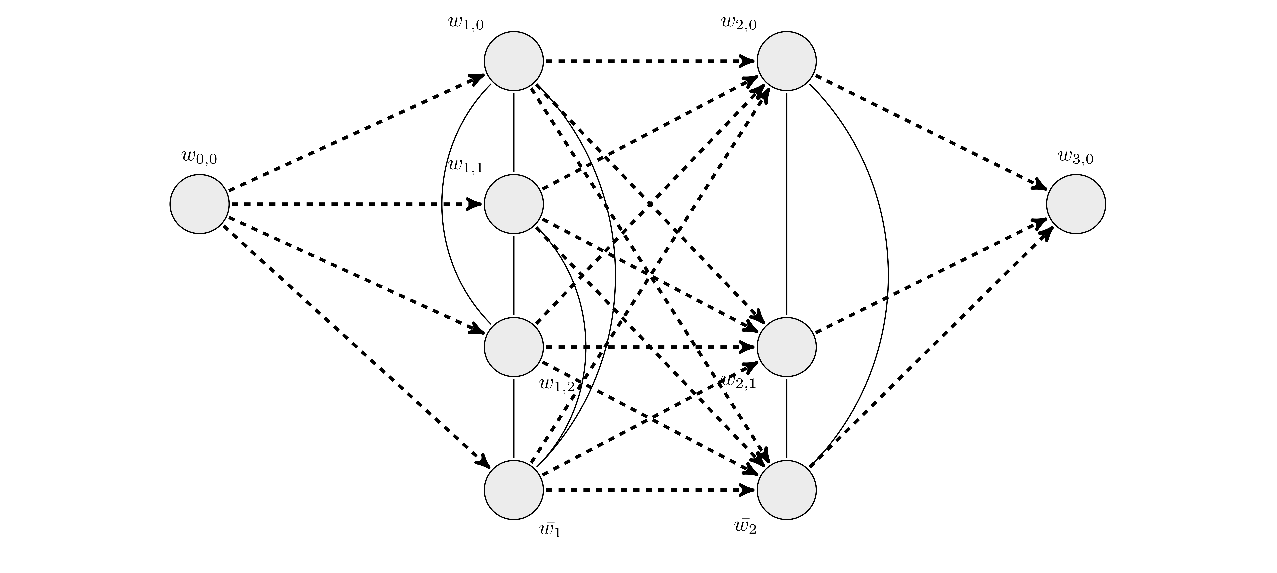}
  \caption{A combination of the epistemic model (see section \ref{sec:PublicAnnouncementLogic}) and the transition system (see section \ref{sec:LTPAL}) for two-framed data. Here, solid lines are epistemic relations, and dotted arrows are transition relations. \\
$M_1 = (W_1, R_1, V_1), W_1 =\{w_{1,0},w_{1,1},w_{1,2},\Bar{w_1}\}, \\R_1 =\{(w_{1,0},w_{1,0}),(w_{1,1},w_{1,1}),(w_{1,2},w_{1,2}),(w_{1,0},w_{1,1}),(w_{1,0},w_{1,2}),(w_{1,1},w_{1,2}),\\(w_{1,0},\Bar{w_1}),(w_{1,1},\Bar{w_1}),(w_{1,2},\Bar{w_1}),(\Bar{w_1},w_{1,0}),(\Bar{w_1},w_{1,1}),(\Bar{w_1},w_{1,2}),(\Bar{w_1},\Bar{w_1})\}$;\\
$M_2 = (W_2, R_2, V_2), W_2 =\{w_{2,0},w_{2,1},\Bar{w_2}\}, \\ R_2 =\{(w_{2,0},w_{2,0}),(w_{2,1},w_{2,1}),(w_{2,1},w_{2,0}),(w_{2,0},w_{2,1}),(w_{2,0},\Bar{w_2}),\\(w_{2,1},\Bar{w_2}),(\Bar{w_2},w_{2,0}),(\Bar{w_2},w_{2,1}),(\Bar{w_2},\Bar{w_2})\}$;\\
$\mathcal{TS} = (S, R, s^0, s^{-1}, \rightarrow, L),S=w_{0,0},  \cup W_1 \cup W_2 \cup w_{3,0},s^0=w_{0,0}, s^{-1}=w_{3,0}, R=R_1 \cup R_2, \\ \rightarrow =\{(w_{0,0},w_{1,0}), (w_{0,0},w_{1,1}), (w_{0,0},w_{1,2}), (w_{1,0},w_{2,0}), (w_{1,0},w_{2,1}), (w_{1,1},w_{2,0}),\\(w_{1,1},w_{2,1}), (w_{1,2},w_{2,0}), (w_{1,2},w_{2,0}), (w_{2,0},w_{3,0}), (w_{2,1},w_{3,0}), (w_{0,0},\Bar{w_1}), (\Bar{w_1},\Bar{w_2}),\\(\Bar{w_1},w_{2,0}),(\Bar{w_1},w_{2,1}),(\Bar{w_2},w_{3,0})\}  $; }
  \label{fig:figA0}
\end{figure*}

Figure \ref{fig:figA0} shows a transition system created for two-frame data. In this transition system, $w_{0,0}$ and $w_{3,0}$ are initial and final worlds, defined for simplicity. In the first layer of the transition system, the corresponding Kripke model $\mathcal{M}$ containing worlds $w_{1,0}$,  $w_{1,1}$, and $w_{1,2}$ represent knowledge extracted by classifiers for the first input data. As it can be seen, classifiers provided three possibilities for output classes of this input data, which are fully connected by epistemic relations (solid lines). These relations demonstrate that these classifiers could not decide which world demonstrates the correct output. Similarly, $M_2$ was created using the second input data of the data stream. Here, dotted lines are transition relations, and the transition system $\mathcal{TS}$ could be defined. This figure provides numbers at the top of each transition relation $w_{i,j}\rightarrow w_{i+1,k}$, which shows the calculated probability of knowledge change between world $w_{i,j}$ and $w_{i+1,k}$.
Herein, a model for two-framed data is illustrated, in which the most probable path is $w_{0,0}w_{1,0}w_{2,0}w_{3,0}$. This path could lead us to correct the misclassifications of classifiers.

\section{Formal verification of Classifiers}
\label{sec:FormalverificationofClassifiers}
Generally, formal verification is a mathematical process to ensure that a model fulfills some property function in an environment  \cite{bjesse2005formal}. Thus, in order to define the verification of classifiers, we should ensure that the defined property function holds for the classifier.   
\begin{definition}
A property set $\rho$ for elements of a set $X$ defines with a property function $f_\rho: X \rightarrow \{ true, false\}$, in which $\rho(X) = \{x \in X \mid f_\rho(x) = true \}$.  
\end{definition}
A single-object classifier is a classifier detecting a single object from an input. In some critical cases, single-object classifiers provide a set of answers because the confidence level of the classification process does not satisfy the minimum requirements. For example, in the YOLO classifier threshold could be defined; and the outputs would be a set of answers with the scores over the threshold (Redmon et al. \cite{redmon2016you}). In another study, a set of inputs was created with a neighborhood and manipulation of the input. In this study, the output set would be all outputs of the classifier for the input set (Huang et al. \cite{Huang0}). In this study, the property set would be denoted by $\rho(X)$; and the set of answers provided by classifier $G$ for input $x$ considering the property $\rho$ would be represented by the notation $[\rho(x)]_G$ . By definition, a single-object classifier should provide a single output class. Thus we define verification as follows:
\begin{definition}
A property  $\rho(x)$ for a single-object classifier $G$ called {\em verified} exactly when  $\mid[\rho(x)]_G\mid=1$.\\
Here, ``$\mid A\mid$'' is the cardinality of the set $A$. 
\end{definition}
In other words, a property for a single-object classifier is verified when it is confident about a single answer (see \cite{HoseinpourDehkordi2020}). 
\begin{definition} 
A property  $\rho(x)$ for a set of single-object classifiers $A_G = \{ G_1, \dots , G_n \}$ called {\em verified} exactly when  $\mid\bigcap_{G \in A_G} [\rho(x)]_G\mid=1$. 
\end{definition}
In a multi-agent scenario, a property for single-object classifiers is verified when multiple classifiers agree about a single answer (see \cite{HoseinpourDehkordi2020}).

Note that we can assume a multi-object classifier as multiple single-object classifiers and the definitions are identical for each single-object classifier. Thus, we use the term classifier instead of the single-object classifier for simplicity.

The definition of verification in a single-frame classifier is based on the appearance of objects in that input frame, but for data-stream, we consider objects in some frames. For example, in a video stream, we could verify the appearance of a dog after a cat. In this case, it is enough to detect an input frame with a dog after seeing a cat's input frame. Thus, we define the property in the form of LTPAL in section  \ref{sec:QALTPAL}. 

\section{Public Announcement Logic}
\label{sec:PublicAnnouncementLogic}
In this section, we are going to introduce a logical model for interpreting single-framed data. 
The epistemic model is based on an extension of PAL that is mentioned in \cite{HoseinpourDehkordi2020}. It will be applied to extract knowledge from single-framed data (i.e., an image). 

Let us first introduce the syntax and semantics of the logic. 
\begin{definition}[Language of $PAL$]
Let $Ag=\{1,\ldots,n\}$ be a finite set of agents. The syntax of the language PAL is as follows, in BNF:
$$\phi::= p\mid\neg\phi\mid(\phi\land\phi)\mid
K_i\phi\mid D_{A}\phi \mid[\phi]\phi.$$
\end{definition}
Where $p$ is a propositional variable (atomic formula) is a pair of form $(x, c)$, in which $x$ denotes the input data and $c$ denotes the target class and $A\subseteq {Ag}$ is a subset of agents. We also notice that $K_i \phi$ read as ``i-th classifier knows $\phi$'' ($i \in$ set of all classifiers). In other words, the i-th classifier was assured about the truth value of $\phi$.
Following that, $D_A \phi$ read as ``$\phi$ is a distributed knowledge in group $A$ of classifiers''. The formula $D_A \phi$ holds exactly when, aggregation of knowledge of agents (equivalently $\cap_{i\in A} R_i$) in the group $A$, satisfies $\phi$ (if $A=\{ i \}$ then $K_i \equiv D_A$). Finally, the formula $[\psi]\phi$ reads as ``after a correct announcement of $\psi$, $\phi$ will hold''. By employing this operator, it could investigate that which knowledge is missing.

Suppose $\mathbb{G} = \{G_1, \dotsc, G_n\}$ is a finite set of classifiers, $X$ is the set of input points, and $C$ is the set of all output classes. A PAL Kripke model is a tuple $M = (W, R_1, \dots , R_n, V)$, where $W$ is a set of worlds (hereabouts, the set $W$ represents all possible output results of the input data), 
$$W := \{c \in C \mid \exists x\in X\ \exists G \in \mathbb{G}\ \text{such that}\ c \in [\rho(x)]_G\}\cup\{\Bar{c}\}.$$
$R_i \subseteq W \times W$ is an equivalent relation between worlds for each classifier $G_i$ in $\mathbb{G}$ .

\begin{equation*}
R_i(c) :=
\begin{cases}
\{c\} & c \notin [\rho(x)]_{G_i} \\
\{c' \mid c' \in [\rho(x)]_{G_i}\}\cup\{\Bar{c}\}\ &  c \in [\rho(x)]_{G_i}
\end{cases}
\end{equation*}

$R_i(\Bar{c})$ are defined as follows:
$$R_i(\Bar{c}) := \{c' \mid c' \in [\eta(x)\cup\mu(x)]_{G_i}\}{\cup\{\Bar{c}\}}, $$

The intended meaning of $w R_i w'$ relation is, the worlds $w$ and $w'$ cannot be epistemically distinguished by the $i$-th classifier. Finally, $V: W \longrightarrow 2^{Prop}$ is the evaluation function specified that the knowledge is represented in any world, where $Prop$ is the set of all atoms. $V(c)$ and $V(\Bar{c})$ are defined as follows:
$$V(c) := \{(x,c) \mid \exists G \in \mathbb{G}\ \text{such that}\ c\in [\rho(x)]_G\}.$$

$$V(\Bar{c}) := \{(x,c) \mid \exists G \in \mathbb{G}\ \text{such that}\ c\in [\rho(x)]_G\}.$$

We extend the evaluation function $V$ to all formulas as follows:
\begin{itemize}
    \item 
$\mathcal{M},w\models p$\quad iff \quad $p\in V(w)$,
\item
$\mathcal{M},w\models \neg\phi$ \quad iff\quad
 $\mathcal{M},w\nvDash\varphi$,
 \item
$\mathcal{M},w\models \phi\land\psi$\quad iff\quad 
$\mathcal{M},w\models\phi\textrm{~and~}\mathcal{M},w\models\psi$,
\item
$\mathcal{M},w\models K_i\phi$\quad iff \quad
$\forall v\in R_i(w),  \mathcal{M},v\models\phi$,
\item
$\mathcal{M},w\models D_{A}\phi$ \quad iff
\quad $\forall v\in R_{A} (w), \mathcal{M},v\models\phi$, where $R_{A}:=\underset{i\in A}{\bigcap} R_i$.  
\item
$\mathcal{M},w\models[\psi]\phi$\quad iff\quad
$\mathcal{M},w\models\psi$ implies  $\mathcal{M}^{\psi},w\models\phi$.
\end{itemize}
Where $\mathcal{M}^\psi:=(W^\psi,R_1^\psi,\ldots, R_n^\psi,V^\psi)$ with

$W^\psi:=\{w\in W; \mathcal{M},w\models\psi\}$,

$R_j^\psi:=R_j\cap (W^\psi\times W^\psi)$ for all $j\in\{1,\ldots,n\}$ and

$V^\psi(w):=V(w)\textrm{~for~all~} w\in W^{\psi}$.

The intended meaning of satisfaction of an atomic formula $(x,c)$ in a world $w$ is ``class $c$ has appeared as respective classifier's output class for the input data $x$''.  We also mention that for every formula $\phi$ and $\psi$, $\phi \vDash \psi$ means that for every model $M$, and every point $w \in M$, $M,w \vDash \phi$ implies that $M,w \vDash \psi$.

Now we can show that a point is verified for classifiers exactly when its interpretation is valid in $\mathcal{M}$.

\begin{theorem}
Suppose that $\mathbb{G} = \{G\}$ is a classifier, $x$ is an input and $c$ is an output class of $G$. Then:
\begin{center}
    $[\rho(x)]_{G} = \{c\}$ exactly when $\mathcal{M}, c \models K p_c$,  where $(x,c)$ denoted by $p_c$.
\end{center}
\end{theorem}

\begin{theorem}
Suppose that  $\mathbb{G} = \{G_1, \dotsc, G_n\}$, is a multi-classifier system such that $\bigcap_{G\in \mathbb{G}}[\rho(x)]_{G} \neq \emptyset$,  $x$ is an input and $c$ is an output class for classifiers in $\mathbb{G}$, then:
\begin{center}
    $\bigcap_{G\in \mathbb{G}}[\rho(x)]_{G} = \{c\}$ exactly when $\mathcal{M}, \Bar{c} \vDash D_{\mathbb{G}}p_c$,  where $(x,c)$ denoted by $p_c$.
\end{center}
\end{theorem}

More details about the model and proofs could be found in Dehkordi et al. \cite{HoseinpourDehkordi2020}.


Here, algorithm \ref{alg:PALS} was developed to investigate the correctness of PAL formulas.  
\begin{theorem}
 Let $\mathcal{M}=(W,R_1, \cdots, R_n, V)$ be a Kripke model, $w\in W$ and $\phi$ an PAL formula. Then, $\mathcal{M},w \vDash \phi$ if and only if PALS($\phi$,$\mathcal{M}$, $w$ ). 
\end{theorem}
\begin{proof}
  We only prove the if direction, the other direction is proved similarly.

``If'' direction: The proof is by induction on the complexity of $\Phi$.
Suppose that $\mathcal{M},w \vDash \phi$, and $\phi \in$PAL.

\begin{itemize}
\item  If $\phi$ is an atomic formula, then the \textit{if condition} at line 3 would be satisfied, so line 4 would be executed. Here, we will check whether $\phi$ is in $V(w)$ or not. By definition we have $\mathcal{M},w\models p$ if and only if  $p\in V(w)$.
    \item If $\phi \equiv \neg \phi_1$, then in this case, the \textit{if condition} at line 7 would be satisfied, so line 8 would be executed. At this line we return the negation of PALS($\phi_1$,$\mathcal{M}$, $w$). And by induction PALS($\phi_1$,$\mathcal{M}$, $w$) if $\mathcal{M},w \vDash \phi_1$.
    \item If $\phi \equiv \phi_1 \land \phi_2$, then in this case, the \textit{if condition} at line 10 would be satisfied, so line 11 would be executed. At this line we return the conjunction of PALS($\phi_1$,$\mathcal{M}$, $w$) and PALS($\phi_2$,$\mathcal{M}$, $w$). And by induction PALS($\phi_1$,$\mathcal{M}$, $w$) if $\mathcal{M},w \vDash \phi_1$ and PALS($\phi_2$,$\mathcal{M}$, $w$) if $\mathcal{M},w \vDash \phi_2$ .
    \item If $\phi \equiv K_i\phi_1$, then in this case, the \textit{if condition} at line 13 would be satisfied, so line 14 would be executed. In this line for every $w'\in R_i(w)$, PALS($\phi_1$,$\mathcal{M}$, $w'$) would be calculated; \textit{true} would be returned if and only if all of them are satisfied. By induction, we have  PALS($\phi_1$,$\mathcal{M}$, $w'$) if $\mathcal{M},w' \vDash \phi_1$. By definition, we have $\forall w'\in R_i(w),  \mathcal{M},w'\models\phi_1$ if $\mathcal{M},w\models K_i\phi_1$. So, $\land_{w'}$PALS($\phi_1, \mathcal{M}, w'$);$\forall w' \in R_i(w)$ if $\mathcal{M},w\models K_i\phi_1$. Consequently, PALS($\phi, \mathcal{M}, w$) if $\mathcal{M},w\models \phi$.
    \item If $\phi \equiv D_A\phi_1$, then in this case, the \textit{if condition} at line 16 would be satisfied, so lines 17 and 18 would be executed. In line 17, first, $R_A$ would be calculated same as $R_{D_A}$ in the definition. Then, for every $w'\in R_A(w)$, PALS($\phi_1$,$\mathcal{M}$, $w'$) would be calculated; \textit{true} would be returned if and only if all of them are satisfied. By induction, we have  PALS($\phi_1$,$\mathcal{M}$, $w'$) if $\mathcal{M},w' \vDash \phi_1$. By definition, we have $\forall w'\in R_{D_A}(w),  \mathcal{M},w'\models\phi_1$ if $\mathcal{M},w\models D_A\phi_1$. So, $\land_{w'}$PALS($\phi_1, \mathcal{M}, w'$);$\forall w' \in R_A(w)$ if $\mathcal{M},w\models D_A\phi_1$. Consequently, PALS($\phi, \mathcal{M}, w$) if $\mathcal{M},w\models \phi$.
    \item If $\phi \equiv [\phi_2]\phi_1$, then in this case, the \textit{if condition} at line 20 would be satisfied, so lines 21 to 25 would be executed. In line 21, first, we check if PALS($\phi_2, \mathcal{M}, w$) is false. In this case, by definition, $\mathcal{M},w\models\phi_2$ implies  $\mathcal{M}^{\phi_2},w\models\phi_1$ would be \textit{true}. Because, by induction we had PALS($\phi_2$,$\mathcal{M}$, $w$) if $\mathcal{M},w \vDash \phi_2$. Otherwise, we could calculate $\mathcal{M}^{\phi_2}$ with the function SFKE($\mathcal{M}, \emptyset, \{\phi_2\}$), in which no input rules should be defined (lemma \ref{lemma:SFKE}). Without input rules, SFKE returns a modification of the Kripke model $\mathcal{M}$, in which all worlds $w$, $\mathcal{M},w \nvDash \phi_2$ are removed. Therefore, SFKE($\mathcal{M}, \emptyset, \{\phi_2\}$)$\equiv \mathcal{M}^{\phi_2}$. By induction, we have  PALS($\phi_2$,$\mathcal{M}$, $w$) if $\mathcal{M},w \vDash \phi_2$ and PALS($\phi_1$,$\mathcal{M}^{\phi_2}$, $w$) if $\mathcal{M}^{\phi_2},w \vDash \phi_1$. By definition, we have $\mathcal{M},w\models\phi_2$ implies  $\mathcal{M}^{\phi_2},w\models\phi_1$ if $\mathcal{M},w\models[\phi_2]\phi_1$. Consequently, PALS($\phi, \mathcal{M}, w$) if $\mathcal{M},w\models \phi$.
\end{itemize}
\end{proof}

\begin{algorithm}[!ht]
Let $\phi$ be the PAL formula, $\mathcal{M}$ the Kripke model, and $w$ a world\;
\caption{The PAL Satisfaction function (PALS) shall investigate the satisfaction of PAL formulas. }\label{alg:PALS}
\begin{algorithmic}[1]
\Function{PALS}{$\phi, \mathcal{M}, w$}
    \\\Comment{$R_i$ and $V$ could be driven from the model $\mathcal{M}$}
    \If{$\phi$ is an atomic formula}
		\State \textbf{return} $\phi \in V(w)$\\ \Comment{This will be true if $\phi \in V(w)$ and false otherwise  }
	\EndIf
	\If{$\phi$ is in form of $\neg \psi$}
	    \State \textbf{return} $\neg$ PALS($\psi, \mathcal{M}, w$)
	\EndIf
	\If{$\phi$ is in form of $\psi_1 \land \psi_2$}
	    \State \textbf{return} PALS($\psi_1, \mathcal{M}, w$)$\land$PALS($\psi_2, \mathcal{M}, w$)
	\EndIf
	\If{$\phi$ is in form of $K_i \psi$}
	    \State \textbf{return} $\land_{w'}$PALS($\psi, \mathcal{M}, w'$);$\forall w' \in R_i(w)$
	\EndIf
	\If{$\phi$ is in form of $D_A \psi$}
	    \State $R_{A} := \underset{i\in A}{\bigcap} R_i$
	    \State \textbf{return} $\land_{w'}$PALS($\psi, \mathcal{M}, w'$);$\forall w'\in R_{A} (w)$
	\EndIf
	\If{$\phi$ is in form of $[\psi_2] \psi_1$}
	    \If{$\neg$ PALS($\psi_2, \mathcal{M}, w $)}
    	    \State \textbf{return} true
	    \Else
    	    \State \textbf{return} PALS($\psi_1$, SFKE($\mathcal{M},\emptyset, \{\psi_2\}$), $w $)
	    \EndIf
	\EndIf
	
	\State \textbf{return} false
\EndFunction
\end{algorithmic}
\end{algorithm}

\section{Linear Temporal Public Announcement Logic}
\label{sec:LTPAL}

In this section, to extract all possible sequences, named execution paths, of knowledge represented in time-series data.

\begin{definition}[Language of LTPAL ]
 We introduce an LTL extension of PAL by the following grammar, in BNF: 
 
$$\phi::= p \mid \neg\phi\mid(\phi\land\phi)\mid
K_j\phi\mid D_{A}\phi \mid[\phi]\phi  $$
$$\Phi::=\phi \mid (\neg\Phi) \mid (\Phi\land\Phi) \mid  X\Phi \mid [\Phi U \Phi] $$
\end{definition}

Here, the temporal operators are $X\Phi$ (in the ne$X$t data-frame $\Phi$ must be true) and $[\Phi U \Psi]$ ($\Psi$ must remain true $U$ntil $\Phi$ becomes true). To define the Kripke semantic of this logic, assume that $M_0=(\{ w_{0_0} \},R_{0_0}, \dots,R_{0_k},V_0) , M_1=(W_1,R_{{1}_0}, \dots, R_{{1}_k},V_{1}), \dots, M_{n-1}=(W_{n-1},R_{{(n-1)}_0}, \dots, R_{{(n-1)}_k},V_{n-1}),\ and\  M_n=(\{ w_{n_0} \},R_{n_0}, \dots,R_{n_k},V_{n})$ are PAL models, where $W_i$'s are mutually disjoint sets, and $V_0(w_{0_0}) = V_n(w_{0_0})  = \emptyset$. 

We build a new model $\mathcal{TS} = (S, R, s^0, s^{-1}, \rightarrow, L)$, known as a transition system, in which $S = \bigcup_{i = 0}^n W_i $ is the set of states, $R =  \{ R_{i_j} \mid 0 \leq i \leq n,\ 0\leq j \leq k \} $, $s^0= w_{0_0}$ and $s^{-1}= w_{n_0}$ are initial and final states, $\rightarrow^k = W_k \times W_{k+1},\  0 \leq i < n$ are transition relations, $\rightarrow=\bigcup_{0 \leq k<n} \rightarrow^k$,  $(w_{ki}, w_{(k+1)j}) \in \rightarrow^k$ is denoted by $\rightarrow_{ij}^k$, and the labeling function $L : S \longrightarrow 2^{Prop}$ which assigning propositional letters to states is defined by $L(w_{i_j}) = V_i(w_{i_j})$, for each  $w_{i_j} \in W_i$. To extend the labeling function to all LTPAL formulas, we first fix the following notations.

An executive path $\pi_\mathcal{TS}^{i\dots j}$ in the model $\mathcal{TS}$ is a sequence of worlds $w_i w_{i+1}\dots w_j$ where 
$w_i \rightarrow w_{i+1} \rightarrow \dots \rightarrow w_{j}$
for $  w_k \in W_k,\  i \leq k \leq j$. An executive path $\pi_\mathcal{TS}^{i\dots j}$ is called total and denoted by $\pi_\mathcal{TS}$, if $w_i = s^0$ and $w_j = s^{-1}$. The set of all total executive paths will be denoted by $\Pi_\mathcal{TS}$. Next, for $\phi \in$ PAL and $\Phi,\Phi_1, \Phi_2 \in $ LTPAL, we define:

\begin{itemize}
    \item 
$\mathcal{TS}, \pi_\mathcal{TS}^{i\dots j} \vDash \phi  $ \quad iff\quad   $M_i, w_i \vDash \phi $,
    \item 
$\mathcal{TS}, \pi_\mathcal{TS}^{i\dots j} \vDash X\Phi  $ \quad iff\quad 
$\mathcal{TS}, \pi_\mathcal{TS}^{i+1\dots j} \vDash \Phi $,
 \item
$\mathcal{TS}, \pi_\mathcal{TS}^{i\dots j} \vDash \Phi_1 U \Phi_2  $ \quad iff\quad  there exists $m$, $i \leq m \leq j$, $\mathcal{TS}, \pi_\mathcal{TS}^{m\dots j} \vDash \Phi_2,$  and for all $k$, $i \leq k < m,$ we have $ \mathcal{TS}, \pi_\mathcal{TS}^{i\dots k} \vDash \Phi_1$.

\end{itemize}
In the first clause above, we notice that we look at the transition system $\mathcal{TS}$ over PAL formulas simply as a Kripke model. Moreover, algorithm \ref{alg:TSTS} is developed to create a transition system from PAL models.  
\begin{algorithm}[!ht]
Let $\mathcal{C}$ be the set of classifiers, $\zeta$ the set of rules, $X$ the time-series input data of size $k$ ($X=[x_1,\dots, x_k]$), and $\eta$ the predefined confident neighborhood.
We also use the MASKS function developed in \cite{HoseinpourDehkordi2020}.
\caption{The Time-Series Transition System (TSTS) function shall create the transition system by the given input information. }\label{alg:TSTS}
\begin{algorithmic}[1]
\Function{TSTS}{$\mathcal{C},\zeta,X, \eta$} 
	\State $k := size(X)$
	\State $S := \{w_0 \}$
	\State $W := \{w_0 \}$
	\State $R_1,\dots,R_n := \{(w_0,w_0) \}$
	\State $s^0 := w_0 $
	\State $s^{-1} := w_0 $
	\State $\rightarrow$ $:= \emptyset $
	\State $V(w_0) := \emptyset $
	\State $L(w_0) := \emptyset $
    \State $\mathcal{M} = (W, R_1, \dots, R_n, V)$
	\State $\mathcal{TS} := (S,R_1,\dots,R_n,s^0,s^{-1},\rightarrow, L)$
	\ForAll{$x_0$ in $X$}
	    \State $\_, \mathcal{M} :=  $MASKS$(\mathcal{C},x_0, \eta)$ 
	    \State $\mathcal{M}' :=$ SFKE($\mathcal{M},\zeta,\emptyset$)\\ 
	    \Comment{Kripke model $\mathcal{M}'=(W',R_1',\dots,R_n',V')$}
	    \State $S := S \cup W'$ 
	    \State $\forall_i R_i := R_i \cup R_i'$ 
	    \ForAll{$w \in W$ and $w' \in W'$}
	        \State $\rightarrow$ $:= \rightarrow \cup$ $\{ (w,w') \} $
	    \EndFor
	    \State $W := W'$ 
	\EndFor
    \State $W := \{w_{k+1} \}$ 
	\State $S := S \cup W$ 
    \State $\forall_i R_i := R_i \cup \{(w_{k+1},w_{k+1}) \}$
	\State $s^{-1} := w_{k+1} $
    \ForAll{$w \in W$ and $w' \in W'$}
        \State $\rightarrow$ $:= \rightarrow \cup$ $\{ (w,w') \} $
    \EndFor
	\State \textbf{return} $\mathcal{TS}$
\EndFunction
\end{algorithmic}
\end{algorithm}

By the function TEMS developed in algorithm \ref{alg:TEMS}, temporal formulas could be investigated. Other temporal operators could be driven from the two operators \textit{next} ($X \Phi$), and \textit{until} ($U\Phi$) in the following way:
\begin{itemize}
    \item $F \Phi \equiv (\top U \Phi)$
    \item $(\Phi R \Psi) \equiv \neg (\neg \Phi U \neg \Psi)$
    \item $(\Phi W \Psi) \equiv  ( \Psi R (\Phi \lor \Psi))$
    \item $G \Phi \equiv (\bot R \Phi)$
\end{itemize}
The intended meaning of \textit{future} ($F\Phi$) is ``eventually $\Phi$ becomes true'', \textit{global} ($G\Phi$) is ``$\Phi$ must remain true forever'', \textit{release} ($\Phi R \Psi$) is ``$\Psi$ remains true until and including when $\Phi$ becomes true, if $\Phi$ never becomes true, $\Psi$ always remains true'', and \textit{weak until} ($\Phi W \Psi$) is ``$\Phi$ has to remain true at least until $\Psi$; if $\Psi$ never holds, $\Phi$ must always remain true'' \cite{hossain2019quantified}, \cite{kwiatkowska2010parallel}. 
\begin{algorithm}[!ht]
Let $\Phi$ be the LTL formula, $\mathcal{TS}$ the transition system, and $w$ a world\;
\caption{The TEMporal Satisfaction function (TEMS) function shall investigate the satisfaction of LTPAL formulas. }\label{alg:TEMS}
\begin{algorithmic}[1]
\Function{TEMS}{$\Phi, \mathcal{TS}, \pi_{TS}^{i\dots j}$}
    \\\Comment{$\rightarrow$ could be driven from the model $\mathcal{TS}$}
    \If{$i > j$}
        \State \textbf{return} false
    \EndIf
    \If{$\Phi$ is a PAL formula}
        \State $\mathcal{M} := $ the $i$-th Kripke model of $\mathcal{TS}$
		\State \textbf{return} PALS($\Phi, \mathcal{M}, w_i$)
		\\\Comment{let $w_i$ as first world in path $\pi_{TS}^{i\dots j}$}
	\EndIf
	\If{$\Phi$ is in form of $\neg \Psi$}
	    \State \textbf{return} $\neg$ TEMS($\Psi, \mathcal{TS}, \pi_{TS}^{i\dots j}$)
	\EndIf
	\If{$\Phi$ is in form of $\Psi_1 \land \Psi_2$}
	    \State \textbf{return} TEMS($\Psi_1, \mathcal{TS}, \pi_{TS}^{i\dots j}$)$\land$TEMS($\Psi_2, \mathcal{TS}, \pi_{TS}^{i\dots j}$)
	\EndIf
	\If{$\Phi$ is in form of $X \Psi$}
	    \State \textbf{return} TEMS($\Psi, \mathcal{TS}, \pi_{TS}^{i+1\dots j}$)
	\EndIf
	\If{$\Phi$ is in form of $\Psi_1 U \Psi_2$}
	    \If{TEMS($\Psi_2, \mathcal{TS}, \pi_{TS}^{i\dots j}$)}
	        \State \textbf{return} true
        \EndIf
	    \While{TEMS($\Psi_1, \mathcal{TS}, \pi_{TS}^{i\dots j}$)}
	        \State $i := i+1$
	        \If{TEMS($\Psi_2, \mathcal{TS}, \pi_{TS}^{i\dots j}$)}
	            \State \textbf{return} true
	        \EndIf
	    \EndWhile
	\EndIf
	
	\State \textbf{return} false
\EndFunction
\end{algorithmic}
\end{algorithm}

\begin{theorem}
 Let $\mathcal{TS} = (S,R_1,\dots,R_n,s^0,s^{-1},\rightarrow, L)$ be a Transition System and $\Phi$ an LTPAL formula. Then for $\mathcal{TS}, \pi_{TS}^{i \dots j} \in \Pi$, we have $\mathcal{TS}, \pi_\mathcal{TS}^{i\dots j} \vDash \Phi  $  iff TEMS($\Phi$,$\mathcal{TS}$, $\pi_{TS}^{i \dots j}$ ).  
\end{theorem}
\vspace*{-5mm}
\begin{proof} 
We only prove the if direction, the other direction is proved similarly. The proof is by induction on the complexity of $\Phi$. 

If direction: For any path formula, in which the length of path is lower than zero, lines 3 and 4 would be executed; and \textit{false} will be returned. For any PAL formula, lines 6 to 8 would be executed. Similar to the semantics of LTPAL, we will check whether $\Phi$ is in $PALS(\Phi, \mathcal{M}, w)$ or not. Here, $\mathcal{M}$ is the i-th Kripke model of the Transition system $\mathcal{TS}$.  

Suppose that $\mathcal{TS}, \pi_\mathcal{TS}^{i\dots j} \vDash \Phi  $, and $\Phi \in$LTPAL is not a PAL formula. So, $\Phi$ would be in form of $\neg \Phi_1, \Phi_1 \land \Phi_2, X\phi_1,$ or $\Phi_1 U \Phi_2$. 
\begin{itemize}
    \item $\Phi \equiv \neg \Phi_1$: In this case, the \textit{if condition} at line 11 would be satisfied, so line 12 would be executed. At this line, we return the negation of TEMS($\Phi_1$,$\mathcal{TS}$, $\pi_{TS}^{i \dots j}$ ). Therefore, by induction, TEMS($\Phi_1$,$\mathcal{TS}$, $\pi_{TS}^{i \dots j}$ ) if $\mathcal{TS}, \pi_\mathcal{TS}^{i\dots j} \vDash \Phi_1$.
    Consequently, TEMS($\Phi$,$\mathcal{TS}$, $\pi_{TS}^{i \dots j}$ ) if $\mathcal{TS}, \pi_\mathcal{TS}^{i\dots j} \vDash \Phi  $.
    \item $\Phi \equiv \Phi_1 \land \Phi_2$: In this case, the \textit{if condition} at line 14 would be satisfied, so line 15 would be executed. At this line we return the conjunction of TEMS($\Phi_1$,$\mathcal{TS}$, $\pi_{TS}^{i \dots j}$ ) and TEMS($\Phi_2$,$\mathcal{TS}$, $\pi_{TS}^{i \dots j}$ ). Therefore, by induction, TEMS($\Phi_1$,$\mathcal{TS}$, $\pi_{TS}^{i \dots j}$ ) if $\mathcal{TS}, \pi_\mathcal{TS}^{i\dots j} \vDash \Phi_1  $ and TEMS($\Phi_2$,$\mathcal{TS}$, $\pi_{TS}^{i \dots j}$ ) if $\mathcal{TS}, \pi_\mathcal{TS}^{i\dots j} \vDash \Phi_2  $. Consequently, TEMS($\Phi$,$\mathcal{TS}$, $\pi_{TS}^{i \dots j}$ ) if $\mathcal{TS}, \pi_\mathcal{TS}^{i\dots j} \vDash \Phi  $.
    \item $\Phi \equiv X\Phi_1$:  In this case, the \textit{if condition} at line 17 would be satisfied, so line 18 would be executed. In this line, TEMS($\Phi_1$,$\mathcal{TS}$, $\pi_{TS}^{i+1 \dots j}$ ) would be calculated; \textit{true} would be returned if and only if for the next state of the path $\pi_{TS}^{i \dots j}$, which is $\pi_{TS}^{i+1 \dots j}$, $\Phi_1$ is true. By induction, we have  TEMS($\Phi_1$,$\mathcal{TS}$, $\pi_{TS}^{i+1 \dots j}$ ) if $\mathcal{TS}, \pi_\mathcal{TS}^{i+1\dots j} \vDash \Phi_1  $. By definition, we have $\mathcal{TS}, \pi_\mathcal{TS}^{i+1\dots j} \vDash \Phi $ if $\mathcal{TS}, \pi_\mathcal{TS}^{i\dots j} \vDash X\Phi  $. Consequently, TEMS($\Phi$,$\mathcal{TS}$, $\pi_{TS}^{i \dots j}$ ) if $\mathcal{TS}, \pi_\mathcal{TS}^{i\dots j} \vDash \Phi  $.
    \item $\Phi \equiv \Phi_1 U \Phi_2$: In this case, the \textit{if condition} at line 20 would be satisfied, so lines 21 to 29 would be executed. In line 21, first, we check if TEMS($\Phi_2$,$\mathcal{TS}$, $\pi_{TS}^{i \dots j}$ ) is true. By definition, we have: $\mathcal{TS}, \pi_\mathcal{TS}^{i\dots j} \vDash \Phi_1 U \Phi_2  $ if  there exists $m$, $i \leq m \leq j$, $\mathcal{TS}, \pi_\mathcal{TS}^{m\dots j} \vDash \Phi_2,$  and for all $k$, $i \leq k < m,$ we have $ \mathcal{TS}, \pi_\mathcal{TS}^{i\dots k} \vDash \Phi_1$. Hence, if $m=i$, TEMS($\Phi$,$\mathcal{TS}$, $\pi_{TS}^{i \dots j}$ ) would be true. Otherwise, by definition TEMS($\Phi_1$,$\mathcal{TS}$, $\pi_{TS}^{i \dots j}$ ) must be true over the path while TEMS($\Phi_2$,$\mathcal{TS}$, $\pi_{TS}^{i \dots j}$ ) satisfied. If TEMS($\Phi_2$,$\mathcal{TS}$, $\pi_{TS}^{i \dots j}$ ) never satisfied while TEMS($\Phi_1$,$\mathcal{TS}$, $\pi_{TS}^{i \dots j}$ ) is true, false should be returned (line 31). By induction we had TEMS($\Phi_1$,$\mathcal{TS}$, $\pi_{TS}^{i \dots j}$ ) if $ \mathcal{TS}, \pi_\mathcal{TS}^{i\dots k} \vDash \Phi_1$ and TEMS($\Phi_2$,$\mathcal{TS}$, $\pi_{TS}^{i \dots j}$ ) if $ \mathcal{TS}, \pi_\mathcal{TS}^{i\dots k} \vDash \Phi_2$.  Consequently, TEMS($\Phi$,$\mathcal{TS}$, $\pi_{TS}^{i \dots j}$ ) if $\mathcal{TS}, \pi_\mathcal{TS}^{i\dots j} \vDash \Phi  $.
\end{itemize}
\end{proof}

\subsection{Formalizing natural language properties in LTPAL}
\label{sec:QALTPAL}
A straightforward interpretation of human language in modal logics (especially in LTL) is one of the most critical strengths of such logics (see \cite{kress2008translating},\cite{dzifcak2009and}, and \cite{nelken1996automatic}). This kind of interpretation could help robots react to human orders \cite{finucane2010ltlmop}. In order to convert the question into a formula,  we first need to extract all possible answers to the question. Then, after converting each possible answer to the LTL formula using the existing -and abovementioned- approaches, we will investigate the satisfaction of each answer. The developed model would lead us a PAL modifications on such LTL formulas. 
To explain that, let $\Phi( p_1, \dots, p_\sigma )$ be an LTL translated formula and $\phi_i,\ 1 \leq i \leq \sigma$ is a PAL formula. Then, $\Phi(\phi_1, \dots, \phi_\sigma)$ is obtained from $\phi$ by substituting each of $p_i$ with $\phi_i$ for all $1 \leq  i \leq \sigma$, respectively. For instance, for $\Phi(p_1, p_2) = G(p_1 U X p_2)$, $\phi_1 = \neg D_A \neg p_1$, and $\phi_2 = K_i p_2$, we have  $\Phi(\phi_1, \phi_2)=G((\neg D_A \neg p_1) U  ( X (K_i p_2)))$, which, for the transition system $\mathcal{TS}$ and the investigated execution path $\pi_\mathcal{TS}$, means that ``$p_1$ should always be a possible answer until in the world which right after, $p_2$ is a robust knowledge for the i-th agent''. By defining a way for introducing the definition of verified and possibility concerning a group of classifiers, answers could be labeled to assure the questioner, ``how reliable is this answer''. This would lead us to design a more proper system for critical applications. Moreover, by capturing missing knowledge, it could be caught which knowledge is mandatory for the system that is not discovered by classifiers or external, and which part of the system could be manipulated to obtain the missing information. 


To define when a formula is verified, possible, or is missing information in a transition system, we first introduce the following notations.
\begin{notation}
Let $\psi, \phi \in $ PAL, $\Phi,\Phi_1,\Phi_2 \in $ LTPAL, $\Phi_1',\Phi_2' \in $ LTPAL/PAL, $O_1 = \{\land, \lor \}, O_2 = \{U, R \}, O = O_1 \cup O_2$, $f(\land)=\lor,f(\lor)=\land$, $f(U)=R,f(R)=U$. For $\Omega \in \{D_A,\neg D_A \neg ,  K_i, \neg K_i \neg, [\psi]D_A, [\psi]\neg D_A \neg, [\psi]K_i, [\psi]\neg K_i \neg \}$ we define $\Omega \Phi$ as follows:

\begin{equation}
  \Omega \Phi =
    \begin{cases}
      \Omega \Phi_1 & \text{iff $\Phi=\Phi_1$},\\
      X \Omega \Phi_1 & \text{iff $\Phi=X\Phi_1$},\\
      X \Omega \neg \Phi_1 & \text{iff $\Phi=\neg X\Phi_1$},\\
      \Omega \Phi_1' O \Omega \Phi_2' & \text{iff $\Phi=\Phi_1' O \Phi_2'$},\\
      \Omega (\neg \Phi_1' f(O_1) \neg  \Phi_2') & \text{iff $\Phi= \neg(\Phi_1' O_1 \Phi_2')$},\\
      \Omega (\neg \Phi_1 f(O_2) \neg  \Phi_2) & \text{iff $\Phi= \neg(\Phi_1 O_2 \Phi_2)$}.
    \end{cases}   
\end{equation}  
 
\end{notation}

Note that, the transition system is created by aggregation of PAL models. In each PAL model, we had a word named $\xoverline[0.7]{w_i}$, in which all atomic formulas in that model are true. By definition of transition relation, there is a path $\pi_\mathcal{\overline{TS}} = s^0\rightarrow\xoverline[0.7]{w_1}\rightarrow\xoverline[0.7]{w_2}\dots \rightarrow\xoverline[0.7]{w_n}\rightarrow s^{-1}$.

Now, for $\Phi \in $ LTPAL, we say that: 

\begin{enumerate}
    \item $\Phi$ is \textit{verified} for the group $A$ of classifiers exactly when,  we have $TS, \pi_\mathcal{\overline{TS}} \vDash D_A \Phi$.
    \item $\Phi$ is \textit{possible} for the group $A$ of classifiers exactly when, we have $TS, \pi_\mathcal{\overline{TS}} \vDash \neg D_A \neg \Phi$.
    \item A PAL formula $\psi$ is called \textit{verified-missing} information for a formula $\Phi$ in group $A$ of classifiers exactly when, we have $TS, \pi_\mathcal{\overline{TS}} \nvDash D_A \Phi$ and $TS, \pi_\mathcal{\overline{TS}} \vDash [\psi]( D_A \Phi)  $.
    \item A PAL formula $\psi$ is called \textit{possible-missing} information for a formula $\Phi$ in group $A$ of classifiers exactly when, we have $TS, \pi_\mathcal{\overline{TS}} \nvDash \neg D_A \neg \Phi$ and $TS, \pi_\mathcal{\overline{TS}} \vDash [\psi](\neg D_A\neg  \Phi)  $.
\end{enumerate}

{\bf Example.}
(Representation of a Video)
In the previous chapter, a PAL logic for representing image data was illustrated. In this example, assume that the given finite video stream contains $n-1$ images, and models of epistemic information extraction for these images are assumed to be $M_1=(W_1,R_{{1}_0}, \dots, R_{{1}_k},V_{1}), \dots,$ $M_{n-1}=(W_{n-1},R_{{(n-1)}_0}, \dots, R_{{(n-1)}_k},V_{n-1})$. By adding two dummy models, $M_0$ and $M_n$, the transition system $\mathcal{TS} = (S, R, s^0,s^{-1}, \rightarrow, L)$ could be created. 

For clarification, we will describe a real-world situation: Assume a medical \textit{clean-room} in an Operating theater, which is monitored by a camera. The recorded video will be fed into a set of classifiers $A$. The room will be cleaned with an air conditioner together with an ultra-violet (UV) LED. As it is known that the UV LED would damage humans, it must be turned off while a person is in the room. On the opposite, the air conditioner should work during the appearance of a person in that room. It also should be stopped in an empty room for saving electricity. Therefore, the first LTPAL formula (here referred to as protocol) is ``shut down UV while any person monitored'', and the second one is ``shout down the air conditioner when no one is in the room''. The classifiers should answer the vital question of ``How robust are the protocols?''.

Let $p$ be ``at least one human exists'' and $q$ be ``the UV is on'', and $r$ be ``air conditioner is on''. We have following translations:
\begin{itemize}
    \item $G (p \Rightarrow X \neg q)$: which is the translation of ``by observation of human, the UV should be shut down''.
    \item $TS, \pi_\mathcal{\overline{TS}} \vDash G (D_A p \Rightarrow X ( D_A \neg q ))$: by the satisfaction of this formula the satisfaction of the property would be verified within group $A$.
    \item $TS, \pi_\mathcal{\overline{TS}} \vDash G (\neg D \neg p \Rightarrow X ( \neg D q ))$: by the satisfaction of this formula, there is a scenario in which the formula holds. It means that this property is possible within group $A$.
    \item $TS, \pi_\mathcal{\overline{TS}} \vDash G (K_i p \Rightarrow X ( K_i \neg q ))$: by the satisfaction of this formula, this property would be robust for the i-th classifier. 
    \item $TS, \pi_\mathcal{\overline{TS}} \vDash G (\neg K_i \neg p \Rightarrow X ( \neg K_i q ))$: by the satisfaction of this formula, the i-th classifier, consider this property as a possible one.
\end{itemize}

Thus, the formula ``$TS, \pi_\mathcal{\overline{TS}} \vDash G (\neg D_A \neg p \Rightarrow X ( D_A \neg q))$'' will ensure that the system enforces UV-let to turn off in humankind's possible existence. Accordingly, this system leads us to investigate whether protocols are followed or not. 

\section{Conclusions and future work}
In order to provide an interpretation for given answers, and to notice reasoning in critical cases, an approach was designed to 
 formalise acquired knowledge by any classification or input rules for single-frame data. Next, by introducing LTPAL, an extension of PAL, the flow of knowledge in time-series data was modeled. This model gives us a representation of acquired knowledge in a transition system structure. This model could examine the satisfaction of properties (LTPAL formulas) in order to solve the verification problem for classifiers. As an application, we explained a question/answering problem scenario, in which questions are properties to be verified. Consequently, we investigated the reliability of answers (i.e., possible or verified).
Moreover, using this model could answer the question: ``which missing knowledge could lead the system to the correct answer?''. Thus, the developed model was suitable for interpreting and verifying multi-classifiers that aim to classify data streams.  

We are going to extend the model to consider probabilities. Then, we can find the most probable scenario for each data stream to assist classifiers. Besides, we are going to define notations to model multi-object classifiers in a unified way. For example, in an input image, multiple objects could be distinguished by their location in the picture. 

\bibliography{ref.bib}

\begin{appendix}
\section{Tool}
\label{sec:tool}
This section will describe the developed tool, which is done by the Python programming language. This tool is aimed to create an LTPAL model for data-stream inputs. To do this, first, we will collect all knowledge gained from classifiers for specific input. This knowledge could be in two forms. The first is ``predicted output classes'' collected by group $A$ of classifiers, known as internal knowledge and in the form of formulas provided by external resources, known as external knowledge. As mentioned, the internal knowledge is provided from a data-stream so that classifiers would provide output classes for each data-frame of the data-stream. Here, for output classes of each data-frame, a Kripke model would be developed based on the provided internal knowledge. 
After that, the model would be enriched using Sub-feature formulas. Consequently, possible worlds of the Kripke model would be calculated. These steps are formally described in \textbf{Remark 1.} of section \ref{sec:PublicAnnouncementLogic}. Thus far, for each data-frame of the data-stream, a Kripke model was provided. Using these Kripke models and adding them in sequential order, we could create a transition system as we described in section \ref{sec:LTPAL}. Now, LTPAL formulas could be investigated over paths of the transition system. Due to the theory provided in section \ref{sec:QALTPAL}, the following statements would be held:
\begin{itemize}
    \item If the LTPAL formula satisfies over all paths of the transition system, this will be a \textit{verified} formula for the respective group $A$ of classifiers.  
    \item If the LTPAL formula satisfies over at least one path of the transition system, this will be a \textit{possible} formula for the respective group $A$ of classifiers.
    \item If group $A$ consists of just one agent $i$, and if the LTPAL formula satisfies over all paths of the transition system, this will be a \textit{verified} (or \textit{robust}) formula for the respective classifier $i$.  
    \item If group $A$ consists of just one agent $i$, and if the LTPAL formula satisfies over at least one path of the transition system, this will be a \textit{possible} formula for the respective classifier $i$.
    \item If just after changing input information, the LTPAL formula satisfies over all paths of the transition system, this information change will be \textit{verified-missing} information for the formula in the respective group $A$ of classifiers.  
    \item If just after changing input information, the LTPAL formula satisfies over at least one path of the transition system, this information change will be \textit{possible-missing} information for the formula in the respective group $A$ of classifiers.  
    \item If group $A$ consists of just one agent $i$, and if just after changing input information, the LTPAL formula satisfies over all paths of the transition system, this information change will be \textit{verified-missing} information for the formula in the respective classifier $i$.  
    \item If just after changing input information, the LTPAL formula satisfies over at least one path of the transition system, this information change will be \textit{possible-missing} information for the formula in the respective classifier $i$.  
\end{itemize}
\subsection{Atomic Formula}
As it is assumed, classifiers have a specified output domain, which means for a group of classifiers $A$, the aggregated knowledge of output domains of them would be assumed as the group's output domain $C$. The output domain defines all possible outcomes of classifiers for every input. Here, a pair $(x,c)$ for each class $c \in C$ from the output domain of a specific input $x$ will be stored as the atomic formula. Besides, all pairs of the input $x$ and sub-features $\{ (x,c_1), \dots , (x,c_k) \}; (x,c) \vDash (x,c_1) \land \dots \land (x,c_k)$ of these output classes would be counted as atomic formulas. So, we created a  \textit{python-class} namely ``AtomicFormula'' with the property ``inputData'', which is referred to the specific data-frame, the property ``name'', which is referred to the name of the output class, and ``fathers'', which referred to what this atomic formula deducted from (i.e., if $(x,c) \vDash (x,c_1) \land \dots \land (x,c_k)$, $(x,c)$ would be the ``fathers'' of $(x,c_i)_{1\leq i \leq k}$ ). We created an instance for each pair of input $x$ and output class $c$, and then we will create the fathers' property using all sub-features. 

\subsection{Kripke Model}
\label{sec:ToolKripkeModel}
In this section, we will describe the development of the Kripke model in the developed tool. To do this, we keep in mind that a PAL Kripke model is a tuple $M = (W, R_1, \dots , R_n, V)$. Therefore, we developed a \textit{python-class} named ``KripkeModel'', which consists of $W$, $R_i;{1\leq i \leq n}$, and $V$ as its properties. To create a proper \textit{python-class} instance following steps would be considered. First, all output classes considering the predefined neighbourhood provided by classifiers for a specific input data-frame would be collected. Second, the intersection of the classes provided by classifiers as the possible output classes would be calculated (see algorithm \ref{alg:MASKS}). The cardinality power-set of such output classes would be the number of possible worlds in this step. Here, the initial Kripke model could be created in the following manner:
\begin{itemize}
    \item Let the cardinality power-set of output classes as $n$,
    \item $W=\{ w_1, ..., w_n \}$,
    \item $R=\{ (w_i, w_j) \mid 1 \leq i,j \leq n \} $,
    \item $V(w_i) = \{(x,c_i)\} \cup \{ (x, c_{i_j}) \mid  (x,c_i) \vDash (x, c_{i_j}) \}$, all corresponding atomic formulas (atomic formulas with the same ``name'') from the power-set of output classes, plus all of their ``fathers'' from the ``AtomicFormula'' class.
\end{itemize}
After creating the Kripke model $\mathcal{M}$, the PAL formula $\varphi$ could be investigated for each world. Herein, for all provided PAL formulas (external knowledge fed to input), and for all worlds $w \in W$, if $\mathcal{M}, w \nvDash \varphi$, the world is considered impossible and will be removed from the Kripke model $\mathcal{M}$ (see algorithm \ref{alg:SFKE}). In this process, we also remove all relations with one end in $w$ and all evaluation function $V(w)$. Therefore, the created Kripke model considered both internal knowledge and provided external knowledge (in the PAL formula).   
\subsection{Transition System}
In this step, a list of Kripke models were provided, each one related to a data-frame. So, to create the transitions system, first, we created a \textit{python-class}, named ``TransitionSystem'', which contains properties $S$, $R$, $s^0$, $s^{-1}$, $\rightarrow$, and $L$, for a transition system $\mathcal{TS}$. This \textit{python-class} contains a \textit{python-method} ``add\_kripke'' that gets a ``KripkeModel'' object as input and appends it to the end of the ``TransitionSystem''. By calling this \textit{python-method} for all Kripke models, the transition system would be created in sequential order. The next step is to investigate the LTPAL formula $\varphi$ in this transition system. To do this, we should investigate ``for all paths does $\varphi$ satisfied?'' to check whether $\varphi$ is a \textit{verified} formula and ``there exists a path in which, $\varphi$ satisfied?'' to check whether $\varphi$ is a \textit{possible} formula. For both objectives, extraction of all paths is needed, so we did it in the ``get\_all\_paths'' \textit{python-method}. By iterating on extracted paths, together with algorithm \ref{alg:TEMS}, we could check whether a formula is \textit{verified} (if it satisfies in all paths) or \textit{possible} (if it satisfies in at least one path).

\section{Extended Data: how to use the LTPAL tool}\label{secA1}
We developed the LTPAL tool to verify the properties of multiple classifiers. Here, properties would appear in the form of LTPAL formulas. Such properties could be investigated for single-framed data (i.e., images) or data-streams (i.e., videos). 
\subsection{Input}
As said, the tool could apply to an ensemble of classifiers. Therefore, one should collect classifiers that are suitable for the purpose. Then, each input (and its neighbourhoods) should be fed into the classifiers. The output would be a list of detected objects for each data-frame, so there is a list of lists of objects for each data-stream. The input should consist:
\begin{itemize}
    \item LTPAL formulas (properties),
    \item PAL formulas,
    \item Classifiers' output class domain,
    \item Number of frames (1 for single data-frame),
    \item Classifiers' ids,
    \item For each classifier, predictions for each frame should be written,
\end{itemize}
Sample inputs are provided on the tool's webpage (``classifiersPredictions.json'' and ``subsetDict.json'' files). 
\subsection{Output}
The expected output is a log file, which involves:
\begin{itemize}
    \item Kripke model of each data-frame,
    \item All execution paths,   
    \item Evaluation of each property over paths,
    \item Define whether each formula is a verified answer or a possible answer,
    \item The transition system,
\end{itemize}   
Sample output file is provided on the tool's webpage (``output\_log\_file.log'' and ``result.json'' files).
\subsection{Execution}
After configuration of the input files, ``MASKS.py'' could be executed using the Python3 compiler (interpreter). 


The file ``classifiersPredictions.json'' contains all LTPAL and PAL formulas, output classes, image accepted overlaps, number of frames, classifiers' ids, and each frame's extracted information for each classifiers. Sample ``classifiersPredictions.json'':
\begin{lstlisting}[language=json,firstnumber=1]
{
  "formulas_LTPAL": [ "X_iU_i(cat,chair)", "X_i&(K_i~&(cat,dog),cat)", "X_ichair" , "X_idog" ],
  "formulas_PAL": [ "&(K_i~&(cat,dog),cat)" ],
  "allClasses": [ "cat", "dog", "chair" ],
  "overlapTresh": 0.5,
  "number_of_frames": 2,
  "classifiers_ids": [ "1", "2" ],
  "1": [
    [{"name": "cat"},{"name": "dog"},{"name": "chair"} ],
    [{"name": "cat"},{"name": "dog"} ]
  ],
  "2": [
    [{"name": "cat"},{"name": "chair"} ],
    [{"name": "cat"} ]
  ]
}
\end{lstlisting}

The file ``result.json'' would be created after the execution. It contains the created transition system. Sample ``result.json'':
\begin{lstlisting}[language=json,firstnumber=1]
{
    "S": [[[0, 0]], [[1, 2], [1, 3]], [[2, 1]], [[3, -1]]], 
    "R": [[[[0, 0], [0, 0]]], [[[1, 2], [1, 2]], [[1, 2], [1, 3]], [[1, 3], [1, 2]], [[1, 3], [1, 3]]], [[[2, 1], [2, 1]]], [[[3, -1], [3, -1]]]], 
    "s_0": 0, 
    "s_1": -1, 
    "Arrow": [[[0, 0], [1, 2]], [[0, 0], [1, 3]], [[1, 2], [2, 1]], [[1, 3], [2, 1]], [[2, 1], [3, -1]]], 
    "L": {"0_0": [], "1_2": [{"name": "cat"}], "1_3": [{"name": "chair"}]}, {"name": "cat"}], "2_1": [{"name": "cat"}], "3_-1": []}
}
\end{lstlisting}

The execution would also provide a file. This file contains information extracted from the inputs. Each frame would provide each classifier's predictions, the common information, and the Kripke model for that frame. Then, all execution paths would be provided. Next, for each input formula, the tool will provide the state of the formula (i.e., verified/possible). After that, the transition system would be provided. Sample ``output\_log\_file.log'':
\begin{lstlisting}[language=json,firstnumber=1]
MASKS started ---------------------------------->
__________Creating Krikpke Model for frame: 0_______________
__________classifiers_prediction_______________
[[{'name': 'cat'}, {'name': 'dog'}, {'name': 'chair'}], [{'name': 'cat'}, {'name': 'chair'}]]
__________arrayOfOutputClasses_______________
[[(cat), (dog), (chair)], [(cat), (chair)]]
__________Intersected arrayOfOutputClasses_______________
[[(cat), (chair)], [(cat), (chair)]]
The input formula: &(K_i~&(cat,dog),cat) removed list of worlds: [1] with names [['chair']] in the kripke Model of frame: 0
__________kripke_______________ for frame number: 0
(
W=[2, 3],
R=[(2, 2), (2, 3), (3, 2), (3, 3)],
V={2: [{"name": "cat"}], 3: [{"name": "chair"}, {"name": "cat"}]})
__________Creating Krikpke Model for frame: 1_______________
__________classifiers_prediction_______________
[[{'name': 'cat'}, {'name': 'dog'}], [{'name': 'cat'}]]
__________arrayOfOutputClasses_______________
[[(cat), (dog)], [(cat)]]
__________Intersected arrayOfOutputClasses_______________
[[(cat)], [(cat)]]
no world removed by PAL formula for frame number: 1
__________kripke_______________ for frame number: 1
(
W=[1],
R=[(1, 1)],
V={1: [{"name": "cat"}]})
__________Paths_______________
[[(0, 0), (1, 2), (2, 1), (3, -1)], [(0, 0), (1, 3), (2, 1), (3, -1)]]
__________Validating Formulas_______________
The input formula: X_iU_i(cat,chair) is --verified-- 
The input formula: X_iU_i(cat,chair) is --possible-- 
The input formula: X_i&(K_i~&(cat,dog),cat) is --verified-- 
The input formula: X_i&(K_i~&(cat,dog),cat) is --possible-- 
The input formula: X_ichair is --possible-- 
The Transition System model is: (
S=[[(0, 0)], [(1, 2), (1, 3)], [(2, 1)], [(3, -1)]],
R=[[((0, 0), (0, 0))], [((1, 2), (1, 2)), ((1, 2), (1, 3)), ((1, 3), (1, 2)), ((1, 3), (1, 3))], [((2, 1), (2, 1))], [((3, -1), (3, -1))]],
s_0=0,
s_1=-1,
Arrow=[((0, 0), (1, 2)), ((0, 0), (1, 3)), ((1, 2), (2, 1)), ((1, 3), (2, 1)), ((2, 1), (3, -1))],
L={(0, 0): [], (1, 2): [{"name": "cat"}], (1, 3): [{"name": "chair"}, {"name": "cat"}], (2, 1): [{"name": "cat"}], (3, -1): []})
MASKS ended <------------------------------------

\end{lstlisting}

The tool can be found on https://github.com/iuwa/LTPAL.

\end{appendix}

\end{document}